\documentclass[sn-mathphys,Numbered]{sn-jnl}

\usepackage{graphicx}%
\usepackage{multirow}%
\usepackage{amsmath,amssymb,amsfonts}%
\usepackage{amsthm}%
\usepackage{mathrsfs}%
\usepackage[title]{appendix}%
\usepackage{xcolor}%
\usepackage{textcomp}%
\usepackage{manyfoot}%
\usepackage{booktabs}%
\usepackage{algorithm}%
\usepackage{algorithmicx}%
\usepackage{algpseudocode}%
\usepackage{listings}%

\theoremstyle{thmstyleone}%
\theoremstyle{thmstyletwo}%

\theoremstyle{thmstylethree}%

\begin{document}

\title[AI Chatbots as Multi-Role Pedagogical Agents: Transforming Engagement in CS Education]{AI Chatbots as Multi-Role Pedagogical Agents: Transforming Engagement in CS Education}


\author*[1]{\fnm{Cassie Chen} \sur{Cao}}\email{ccao5@sheffield.ac.uk}

\author[2]{\fnm{Zijian} \sur{Ding}}\email{t-zijianding@microsoft,com}

\author[3]{\fnm{Jionghao} \sur{Lin}}\email{jionghal@andrew.cmu.edu}

\author[4]{\fnm{Frank} \sur{Hopfgartner}}\email{hopfgartner@uni-koblenz.de}

\affil*[1]{\orgname{University of Sheffield}, \state{Sheffield}, \country{UK}}

\affil[2]{\orgname{Microsoft Research}, \state{Redmond}, \country{USA}}

\affil[3]{\orgname{Carnegie Mellon University}, \state{Pittsburgh}, \country{USA}}

\affil[4]{\orgname{Universität Koblenz}, \state{Koblenz}, \country{GE}}




\maketitle

\section{Introduction}\label{sec1}

This research is propelled by key challenges in computer science (CS) education, which include low engagement rates, high student-to-teacher ratios, and students' hesitation to engage in inquiry-based learning. These issues, compounded by the abstract nature of CS subject matter and limited opportunities for personalized attention and feedback, create barriers to effective learning.

Conversely, advancements in technology-enhanced learning and the integration of Artificial Intelligence (AI) have ushered in new potential solutions. Research indicates that conversational technology can significantly improve students' learning motivation and performance \cite{yin2021conversation}. Specifically, AI-powered, multi-role chatbots present a promising intervention to personalize learning experiences, foster engagement, and facilitate inquiry-based learning \cite{winkler2018unleashing}.

To capitalize on this potential, our research proposes a learning environment enriched with four distinct chatbot roles: Instructor Bot, Social Companion Bot, Career Advising Bot, and Emotional Supporter Bot. These roles align with the principles of the Self-Determination Theory \cite{deci2013intrinsic}, which suggests that the fulfillment of three innate psychological needs – competence, autonomy, and relatedness – enhances engagement and motivation. For example, the Instructor Bot fosters competence through academic guidance, the Career Advising Bot promotes autonomy by providing personalized career advice, and the Social Companion Bot and Emotional Supporter Bot contribute to a sense of relatedness by offering social and emotional support.

Notably, chatbots are not only supportive tools; when powered by advanced language models like GPT-3, they can drive pedagogical innovation. Recent work has demonstrated the potential of GPT-3-driven pedagogical agents in training children's curious question-asking skills \cite{abdelghani2023gpt}. However, the quality of chatbot interactions is crucial for their acceptance and impact on learning \cite{radziwill2017evaluating}. Therefore, our research aims to optimize the design and role allocation of multi-role chatbots, ensuring their effectiveness in fostering a stimulating and supportive learning environment.

\section{Research Questions}

\begin{enumerate}
    \item How does the use of the multi-role chatbots influence the quality and quantity of student inquiries?
    \item How do the multi-role chatbots affect students’ learning experience, engagement and motivation?
    \item How do students perceive the individual and collective roles of the chatbots?
\end{enumerate}

\section{System Architecture}\label{sec2}

\begin{figure}[h]
\centering \includegraphics[width=1\textwidth]{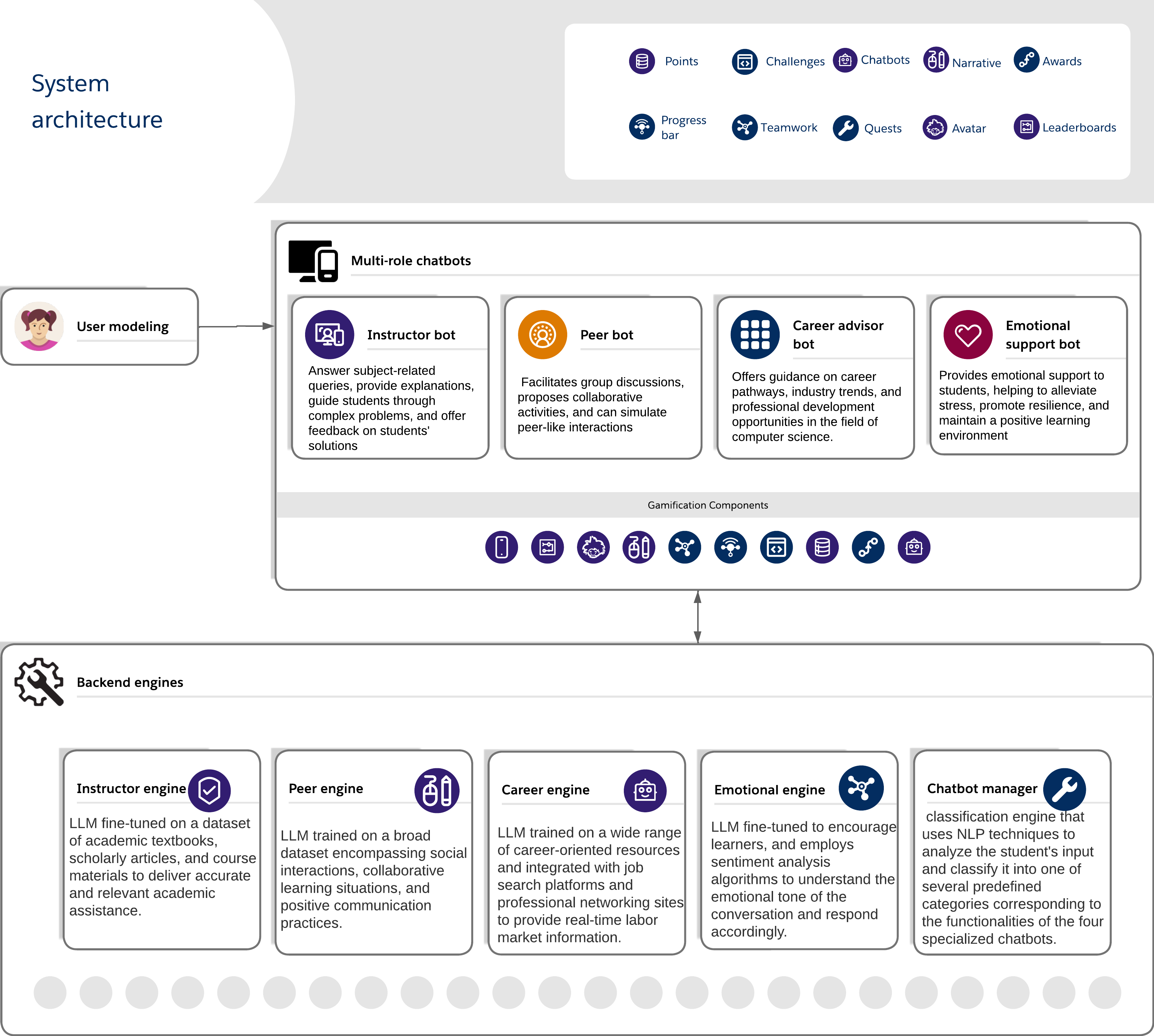}
\caption{System architecture design of 4 distinct bots.}
\label{fig:system}
\end{figure}

The system comprises a full-stack development structure, inclusive of frontend and backend components. The frontend presents users with four distinct chatbots, each bearing unique functionalities as shown in Figure \ref{fig:system}.

The \textbf{Instructor Bot} is designed to offer personalized academic instruction and real-time feedback. This bot caters to the pedagogical needs of students, mitigating the complexity traditionally associated with computer science courses by providing detailed, step-by-step explanations and offering clarification on difficult concepts.

The \textbf{Peer Bot} is engineered to foster a social learning environment by encouraging peer-to-peer interactions. By facilitating collaborative problem-solving and discussions, this bot augments the sense of a vibrant, engaging, and inclusive learning community.

Simultaneously, the \textbf{Career Advising Bot} focuses on bridging the gap between theoretical knowledge and its practical applications in the real world. By providing course-specific career advice and illustrating how curriculum content can translate into potential career paths, this bot serves to contextualize learning within a broader professional scope. In doing so, it enhances student motivation by illuminating the practical relevance and applicability of their coursework.

Finally, the \textbf{Emotional Supporter Bot} recognizes and addresses students' emotional states. It is programmed to provide personalized emotional support and motivational prompts, aiming to alleviate the academic stress and anxiety often experienced by students. By offering such support, this bot aims to foster resilience and sustain persistence in the face of academic adversity.

Moreover, a \textbf{Chatbot Manager} is incorporated to coordinate the operation of all chatbots. It detects the type and complexity of the student's inquiry and decides which chatbot is best equipped to address it, thereby ensuring efficient and effective query resolution.

This architecture facilitates an integrated, responsive, and adaptive learning environment, catering to diverse student needs and fostering a comprehensive and engaging learning experience.

\section{Research Methods}\label{sec3}

This research delves into the potential of multi-faceted, AI-driven chatbots in delivering tailored learning experiences, enhancing student engagement, and facilitating Inquiry-Based Learning. The study leverages a Design-Based Research (DBR) approach \cite{dede2004design}, which encompasses the iterative process of system conception, prototyping, deployment, evaluation, and refinement within a chatbot-enhanced learning environment.

This investigation employs an ABC experimental design, encompassing three distinct conditions: instruction led by a human tutor, instruction directed by a singular chatbot, and instruction steered by a multi-role chatbot.

The study's participants comprise approximately 200 postgraduate students from a Computer Science program at a reputable higher education institution in Germany. These participants are randomly allocated to the three conditions to ensure an impartial distribution.

The research spans a period of one month, throughout which an array of data is garnered. This data encompasses not only the quantity and subject matter of student inquiries but also the qualitative characteristics of interactions between students and their assigned learning facilitators. All data is logged and anonymized to uphold privacy standards.

After the study period, a comprehensive survey is administered to the participants to gather their feedback on the learning experience. This survey is designed to collect both quantitative and qualitative data regarding students' perception of their learning engagement, comprehension of the course material, and their perceived effectiveness of the teaching facilitator.

For further qualitative insights, focus group interviews are conducted. These interviews are transcribed and subject to thematic analysis to identify patterns and outliers in students' experiences.

The study employs Natural Language Processing (NLP) techniques to enhance the analysis of interaction data and survey responses. This includes topic modeling to identify the main topics in the students' questions and discussions. Sentiment analysis is used to infer the emotional state of students from their text input, providing an additional layer to evaluate engagement and motivation. Furthermore, sequence analysis will be employed to study the temporal patterns of interactions between students and chatbots.

Ethical considerations, including data privacy, fairness, and the role of AI in decision-making, are critically examined. Based on empirical findings, the study offers practical recommendations for educators, AI developers, and policymakers to responsibly leverage AI chatbots in optimizing the learning experience in data science courses.

\section{Preliminary Results}\label{sec4}

\begin{figure}[h]
\centering \includegraphics[width=1\textwidth]{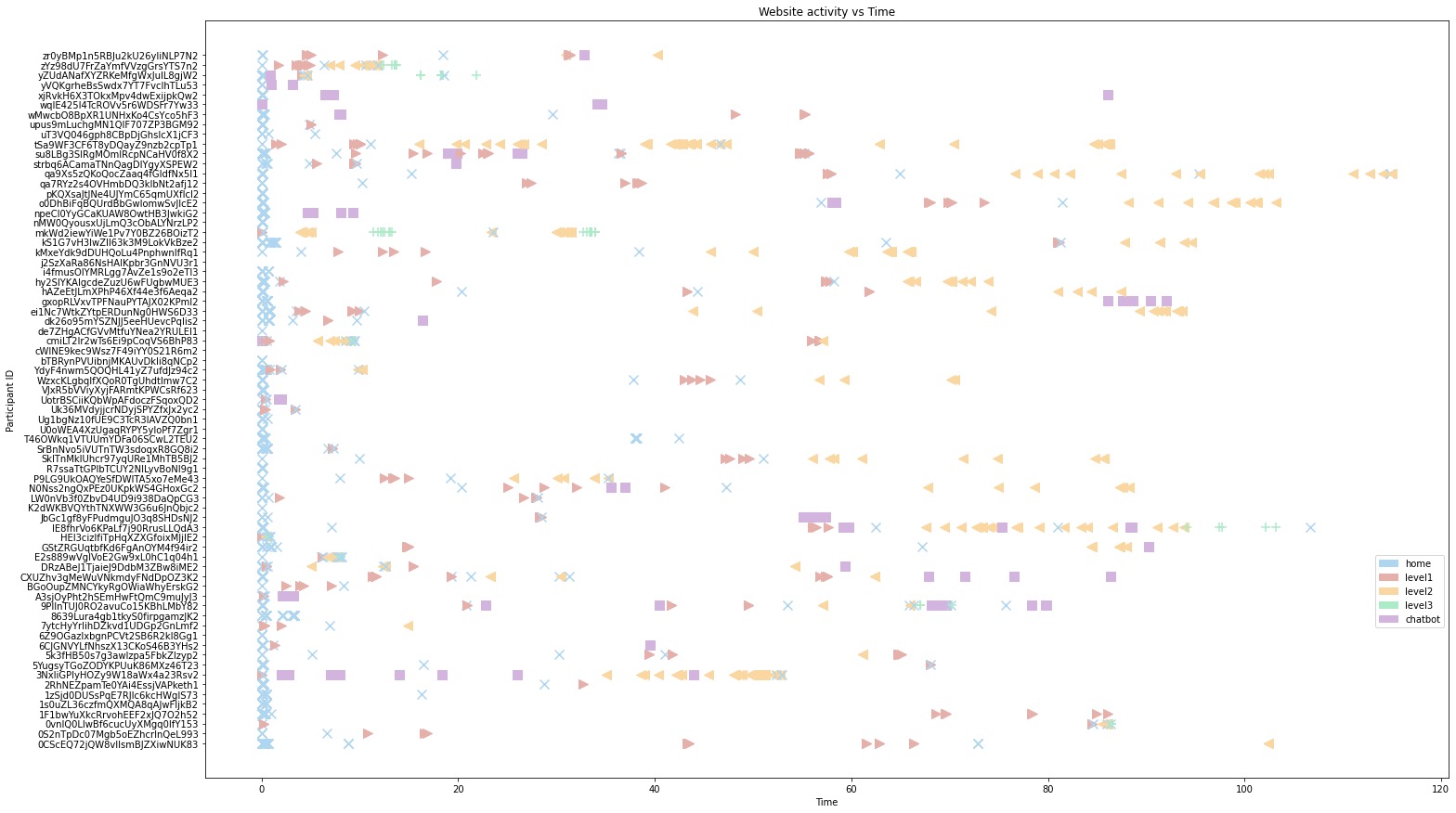}
\caption{Sequence analysis of users’ interaction with the system. The vertical axis represents the users, while the horizontal axis represents the clicks from the users on different pages. The clicks on the homepages were designated as blue squares, the clicks on the level-1 page in the first episode were symbolised as red triangles, the clicks on the level-2 page in the second episode were illustrated as yellow triangles, the clicks on the level-3 page in the final episode were symbolised as green crosses, and the clicks on the chatbot page were represented by purple squares.}
\label{fig:sequence}
\end{figure}

A pilot study was conducted to evaluate the performance of the chatbots as shown in Figure \ref{fig:sequence}. Following Bloom's taxonomy, we generated a dataset combining students’ questions in different cognitive levels and the chatbots’ corresponding answers through machine learning techniques. To evaluate the quality of LLM-generated responses, we developed a measurement metric with five dimensions: accuracy, fluency, empathy, engagement, and relevance (see Table \ref{tab:measurement}).

\begin{table}
    \begin{tabular}{|l|l|l|l|l|l|}
        \hline & \textbf{Definition} & \textbf{Q \& A} & \textbf{Code Check \& Explanation}  \\ \hline
        \textit{Accuracy} & How closely the   & Expert Screening & Ground Truth Checking  \\ 
        &responses match &&\\
        &correct answers&&\\\hline
        \textit{Fluency} & How natural and  & Expert Screening & Expert Screening \\ 
        &smooth the responses & \& Automated &\& Automated Readability\\
        &sound&Readability Index& Index\\\hline
        \textit{Empathy} & How well the chatbot  & Researcher Coding  & Researcher Coding \&\\
        &is able to understand &\& Correlation of&Sentiment Analysis\\
        &and respond to the&Sentiment Analysis&\\
        &emotions of the user&&\\\hline
        \textit{Engagement} & How well the chatbot is & Researcher Coding & Researcher Coding \\ &able to keep the user  & \& Word count &\\
        &interested \& engaged &&\\
        &in the conversation&&\\\hline
        \textit{Relevance} & How closely the & Semantic & [Covered by Accuracy given \\
        &responses are related&Similarity \& Topic &the Ground Truth of Code\\
        &to the topic of &Modelling& Answers]\\
        &the conversation&&\\\hline
    \end{tabular}
    \caption{Measurement metric of five dimensions of GPT generated Q\&A and code check \& explanation.}
    \label{tab:measurement}
\end{table}

Our results demonstrate that LLM is capable of providing accurate and human-like responses to student questions. The chatbot also showed higher engagement and empathy for questions at higher cognitive levels than for questions at lower cognitive levels. Additionally, our semantic similarity analysis and keyword co-occurrence analysis indicate that the LLM-generated responses were relevant and coherent to the student's questions.

Overall, the preliminary findings suggest that large language model-based chatbots can be used as a valuable tool to assist students in their CS1 course, opening up new possibilities to improve the learning experience.

\section{Conclusion}\label{sec13}

Through this exploration, the study contributes to the literature on AI-enhanced learning environments and computer science education. The investigation of the diverse roles of AI chatbots in educational settings provides a timely and practical lens into the transformative potential of AI in fostering engaging, supportive, and personalized learning experiences.

\bibliography{sn-bibliography}


\begin{thebibliography}{6}
\ifx \bisbn   \undefined \def \bisbn  #1{ISBN #1}\fi
\ifx \binits  \undefined \def \binits#1{#1}\fi
\ifx \bauthor  \undefined \def \bauthor#1{#1}\fi
\ifx \batitle  \undefined \def \batitle#1{#1}\fi
\ifx \bjtitle  \undefined \def \bjtitle#1{#1}\fi
\ifx \bvolume  \undefined \def \bvolume#1{\textbf{#1}}\fi
\ifx \byear  \undefined \def \byear#1{#1}\fi
\ifx \bissue  \undefined \def \bissue#1{#1}\fi
\ifx \bfpage  \undefined \def \bfpage#1{#1}\fi
\ifx \blpage  \undefined \def \blpage #1{#1}\fi
\ifx \burl  \undefined \def \burl#1{\textsf{#1}}\fi
\ifx \doiurl  \undefined \def \doiurl#1{\url{https://doi.org/#1}}\fi
\ifx \betal  \undefined \def \betal{\textit{et al.}}\fi
\ifx \binstitute  \undefined \def \binstitute#1{#1}\fi
\ifx \binstitutionaled  \undefined \def \binstitutionaled#1{#1}\fi
\ifx \bctitle  \undefined \def \bctitle#1{#1}\fi
\ifx \beditor  \undefined \def \beditor#1{#1}\fi
\ifx \bpublisher  \undefined \def \bpublisher#1{#1}\fi
\ifx \bbtitle  \undefined \def \bbtitle#1{#1}\fi
\ifx \bedition  \undefined \def \bedition#1{#1}\fi
\ifx \bseriesno  \undefined \def \bseriesno#1{#1}\fi
\ifx \blocation  \undefined \def \blocation#1{#1}\fi
\ifx \bsertitle  \undefined \def \bsertitle#1{#1}\fi
\ifx \bsnm \undefined \def \bsnm#1{#1}\fi
\ifx \bsuffix \undefined \def \bsuffix#1{#1}\fi
\ifx \bparticle \undefined \def \bparticle#1{#1}\fi
\ifx \barticle \undefined \def \barticle#1{#1}\fi
\bibcommenthead
\ifx \bconfdate \undefined \def \bconfdate #1{#1}\fi
\ifx \botherref \undefined \def \botherref #1{#1}\fi
\ifx \url \undefined \def \url#1{\textsf{#1}}\fi
\ifx \bchapter \undefined \def \bchapter#1{#1}\fi
\ifx \bbook \undefined \def \bbook#1{#1}\fi
\ifx \bcomment \undefined \def \bcomment#1{#1}\fi
\ifx \oauthor \undefined \def \oauthor#1{#1}\fi
\ifx \citeauthoryear \undefined \def \citeauthoryear#1{#1}\fi
\ifx \endbibitem  \undefined \def \endbibitem {}\fi
\ifx \bconflocation  \undefined \def \bconflocation#1{#1}\fi
\ifx \arxivurl  \undefined \def \arxivurl#1{\textsf{#1}}\fi
\csname PreBibitemsHook\endcsname

\bibitem[\protect\citeauthoryear{Yin et~al.}{2021}]{yin2021conversation}
\begin{barticle}
\bauthor{\bsnm{Yin}, \binits{J.}},
\bauthor{\bsnm{Goh}, \binits{T.-T.}},
\bauthor{\bsnm{Yang}, \binits{B.}},
\bauthor{\bsnm{Xiaobin}, \binits{Y.}}:
\batitle{Conversation technology with micro-learning: The impact of
  chatbot-based learning on students’ learning motivation and performance}.
\bjtitle{Journal of Educational Computing Research}
\bvolume{59}(\bissue{1}),
\bfpage{154}--\blpage{177}
(\byear{2021})
\end{barticle}
\endbibitem

\bibitem[\protect\citeauthoryear{Winkler and
  S{\"o}llner}{2018}]{winkler2018unleashing}
\begin{bchapter}
\bauthor{\bsnm{Winkler}, \binits{R.}},
\bauthor{\bsnm{S{\"o}llner}, \binits{M.}}:
\bctitle{Unleashing the potential of chatbots in education: A state-of-the-art
  analysis}.
In: \bbtitle{Academy of Management Proceedings},
vol. \bseriesno{2018},
p. \bfpage{15903}
(\byear{2018}).
\bcomment{Academy of Management Briarcliff Manor, NY 10510}
\end{bchapter}
\endbibitem

\bibitem[\protect\citeauthoryear{Deci and Ryan}{2013}]{deci2013intrinsic}
\begin{bbook}
\bauthor{\bsnm{Deci}, \binits{E.L.}},
\bauthor{\bsnm{Ryan}, \binits{R.M.}}:
\bbtitle{Intrinsic Motivation and Self-determination in Human Behavior}.
\bpublisher{Springer}, \blocation{???}
(\byear{2013})
\end{bbook}
\endbibitem

\bibitem[\protect\citeauthoryear{Abdelghani et~al.}{2023}]{abdelghani2023gpt}
\begin{botherref}
\oauthor{\bsnm{Abdelghani}, \binits{R.}},
\oauthor{\bsnm{Wang}, \binits{Y.-H.}},
\oauthor{\bsnm{Yuan}, \binits{X.}},
\oauthor{\bsnm{Wang}, \binits{T.}},
\oauthor{\bsnm{Lucas}, \binits{P.}},
\oauthor{\bsnm{Sauz{\'e}on}, \binits{H.}},
\oauthor{\bsnm{Oudeyer}, \binits{P.-Y.}}:
Gpt-3-driven pedagogical agents to train children’s curious question-asking
  skills.
International Journal of Artificial Intelligence in Education,
1--36
(2023)
\end{botherref}
\endbibitem

\bibitem[\protect\citeauthoryear{Radziwill and
  Benton}{2017}]{radziwill2017evaluating}
\begin{botherref}
\oauthor{\bsnm{Radziwill}, \binits{N.M.}},
\oauthor{\bsnm{Benton}, \binits{M.C.}}:
Evaluating quality of chatbots and intelligent conversational agents.
arXiv preprint arXiv:1704.04579
(2017)
\end{botherref}
\endbibitem

\bibitem[\protect\citeauthoryear{Dede et~al.}{2004}]{dede2004design}
\begin{bchapter}
\bauthor{\bsnm{Dede}, \binits{C.}},
\bauthor{\bsnm{Nelson}, \binits{B.}},
\bauthor{\bsnm{Ketelhut}, \binits{D.J.}},
\bauthor{\bsnm{Clarke}, \binits{J.}},
\bauthor{\bsnm{Bowman}, \binits{C.}}:
\bctitle{Design-based research strategies for studying situated learning in a
  multi-user virtual environment}.
In: \bbtitle{Proceedings of the Sixth International Conference on the Learning
  Sciences},
pp. \bfpage{158}--\blpage{165}
(\byear{2004})
\end{bchapter}
\endbibitem

\end{thebibliography}

\appendix

\newpage

\textbf{Appendix}

\begin{figure}[h]
\centering \includegraphics[width=1\textwidth]{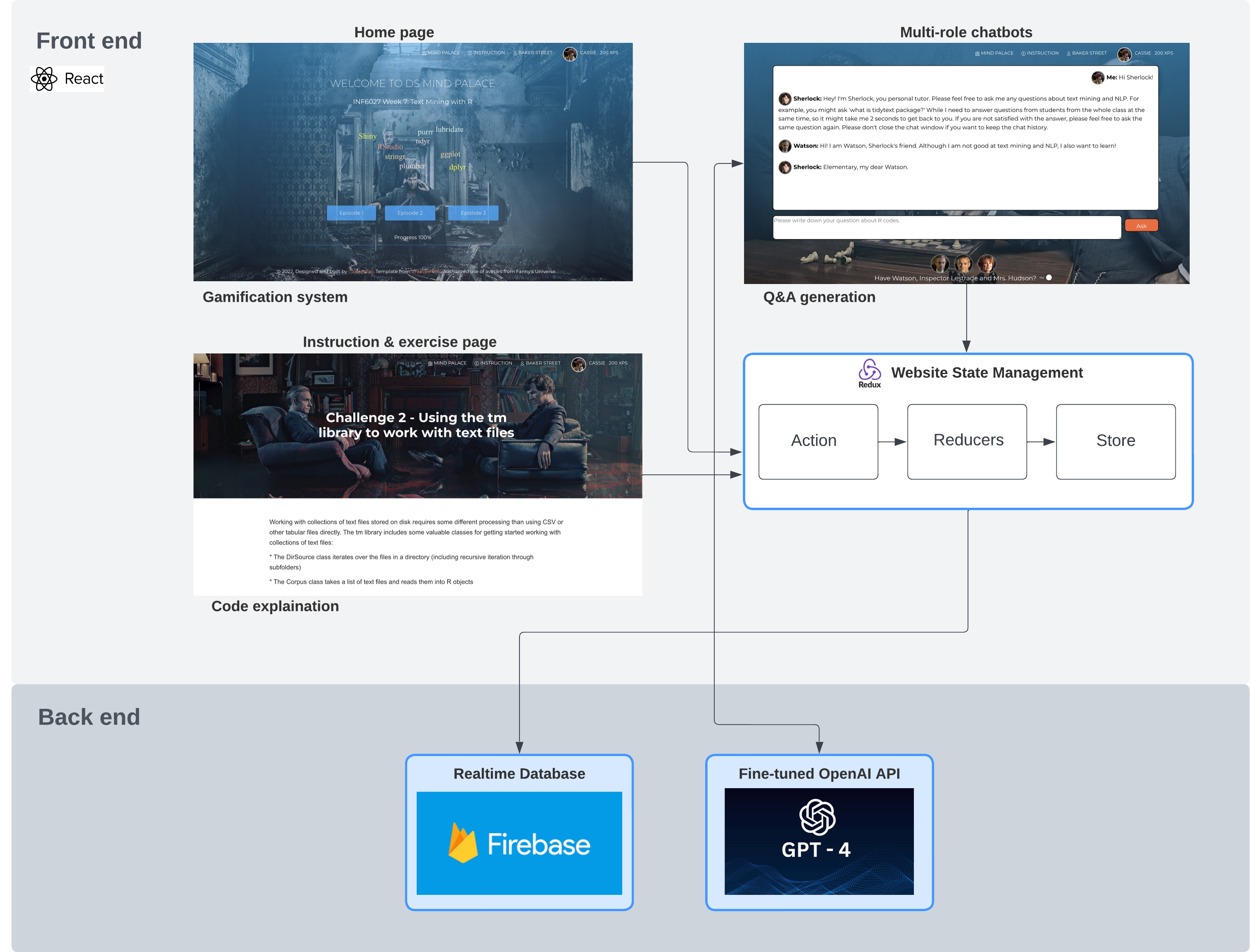}
\caption{Site map of DS Mind Palace gamification learning system.}
\label{fig:flow}
\end{figure}

\begin{figure}[h]
\centering \includegraphics[width=1\textwidth]{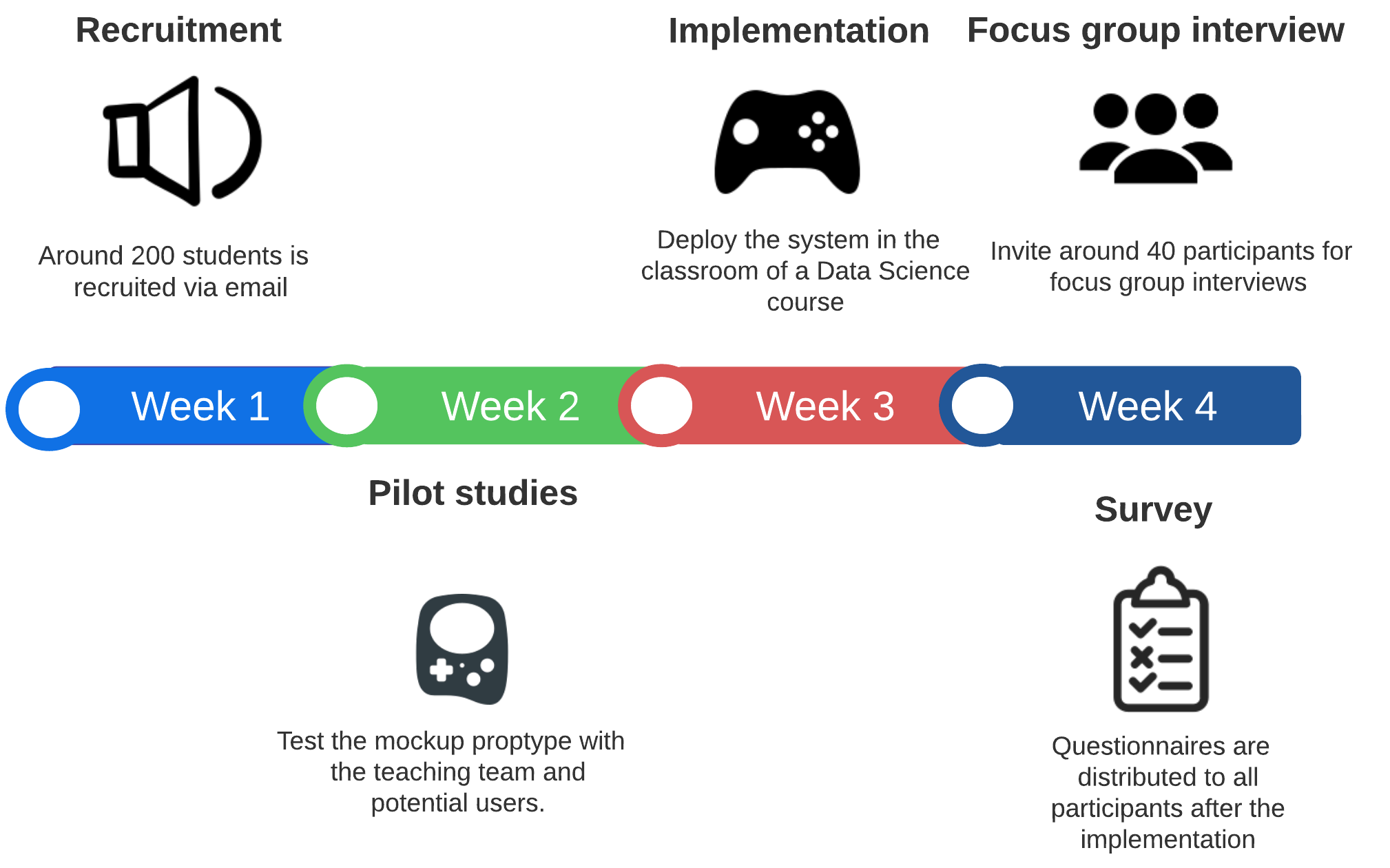}
\caption{Timeline of user study with four stages.}
\label{fig:flow}
\end{figure}

\end{document}